Article

# Explainable Ensemble-Based Machine Learning Models for Detecting the Presence of Cirrhosis in Hepatitis C Patients

**Abrar Alotaibi, Lujain Alnajrani, Nawal Alsheikh, Alhatoon Alanazy, Salam Alshammasi, Meshael Almusairii, Shoog Alrassan and Aisha Alansari ***

College of Computer Science and Information Technology, Imam Abdulrahman Bin Faisal University, P.O. Box 1982, Dammam 31441, Saudi Arabia
* Correspondence: 2180004329@iau.edu.sa

**Abstract:** Hepatitis C is a liver infection caused by a virus, which results in mild to severe inflammation of the liver. Over many years, hepatitis C gradually damages the liver, often leading to permanent scarring, known as cirrhosis. Patients sometimes have moderate or no symptoms of liver illness for decades before developing cirrhosis. Cirrhosis typically worsens to the point of liver failure. Patients with cirrhosis may also experience brain and nerve system damage, as well as gastrointestinal hemorrhage. Treatment for cirrhosis focuses on preventing further progression of the disease. Detecting cirrhosis earlier is therefore crucial for avoiding complications. Machine learning (ML) has been shown to be effective at providing precise and accurate information for use in diagnosing several diseases. Despite this, no studies have so far used ML to detect cirrhosis in patients with hepatitis C. This study obtained a dataset consisting of 28 attributes of 2038 Egyptian patients from the ML Repository of the University of California at Irvine. Four ML algorithms were trained on the dataset to diagnose cirrhosis in hepatitis C patients: a Random Forest, a Gradient Boosting Machine, an Extreme Gradient Boosting, and an Extra Trees model. The Extra Trees model outperformed the other models achieving an accuracy of 96.92%, a recall of 94.00%, a precision of 99.81%, and an area under the receiver operating characteristic curve of 96% using only 16 of the 28 features.

## 1. Introduction

Hepatitis C is a blood-borne virus-related infection triggered by the hepatitis C virus (HCV) that mostly damages the liver. HCV infections are a significant cause of liver diseases, including cirrhosis and hepatocellular carcinoma, making it an important public health concern globally. HCV infections are a prominent cause of liver disease, cirrhosis, and hepatocellular cancer around the globe. As stated by the World Health Organization, 58 million individuals worldwide are infected with hepatitis C at its extreme stages, representing 3% of the global population, and that 1.5 million new cases occur annually. In 2019, almost 290,000 infected patients died from cirrhosis or liver cancer [1]. A 2018 meta-analysis estimated that HCV antibody prevalence in Egypt was 11.9%, making it the country with the highest HCV prevalence worldwide [2]. Therefore, HCV is estimated to be a leading public health dilemma worldwide, which must be addressed with solid program interventions.

Currently, hepatitis C can be cured with direct-acting antiviral (DAA). Nevertheless, many patients still have chronic hepatitis C infection and run the danger of experiencing its associated complications, including cirrhosis and liver cancer. Furthermore, obstacles still stand in the way of attaining widespread access to these medications, particularly in low- and middle-income nations where the hepatitis C epidemic is most severe. According to a study by Elgharably et al. [3], although recent medicines have shown an efficiency rate greater than 90%, access to these new medications is significantly restricted by cost. Additionally, extensive DAA therapy would not eliminate all the issues caused by the HCV pandemic in Egypt. Hepatocellular carcinoma and the sequelae of decompensated cirrhosis continue to place a significant burden on Egyptian society and require sufficient allocation of healthcare resources.

Liver cirrhosis is an advanced stage of liver fibrosis that develops due to the long-term effects of various chronic liver diseases. This condition is characterized by transforming healthy liver tissue into abnormal lesions, accompanied by tissue fibrosis, which can be detected through a medical examination [4]. In the initial stages of cirrhosis, most patients do not experience noticeable symptoms, and the disease is frequently discovered by accident during routine medical examinations for other conditions [5]. When a healthy liver experiences ongoing fibrosis and scarring, intrahepatic resistance in the liver increases, and portal hypertension develops, leading to decreased liver function and fatal consequences [6]. Sepanlou et al. [4] found that, in 2017, cirrhosis was responsible for over 1.32 million fatalities worldwide, with 33.3% of these deaths occurring among females and 66.7% among males. There was a 1.9% mortality rate from cirrhosis in 1990, which has been increasing ever since, reaching 2.4% in 2017. Furthermore, according to the study, Egypt had the highest age-standardized death rate due to cirrhosis, which has remained consistently high since 1990.

Currently, the most accurate way to diagnose cirrhosis is by liver biopsy. However, this is expensive, invasive, and risky. Less invasive tests, such as transient elastography, have therefore been introduced. Transient elastography is a good predictor of HCV and its stages, offering a sensitivity between 72–84% and a specificity between 82–95%. However, it is not widely available [6]. As an alternative to imaging techniques, there are several simple and non-invasive clinical laboratory tests, such as the aspartate aminotransferase-to-platelet ratio index (APRI), aspartate aminotransferase (AST)/alanine aminotransferase (ALT) ratio, alkaline phosphatase, the Naples prognostic score, the Lok score, and the fibrosis index. Those tests have proven to help assess cirrhosis or fibrosis. Yet, they suffer from some limitations and can be used only for certain cases [6–8]. These limitations include unpredictability, insufficient accuracy, and risk factors for error. Additionally, the development of new biomarkers for fibrosis may be constrained by the inherent sample error associated with the current reference standard [9]. Furthermore, FibroTest is a tool that has been widely used for assessing liver fibrosis. Nevertheless, FibroTest is a static examination that offers a single score instantly. As a result, it would not be able to record changes in liver fibrosis over time, which might be crucial for tracking the development of the illness and directing treatment choices [10]. Moreover, due to technical restrictions, FibroTest cannot be used on all patients, including those with ascites, those who are morbidly obese, and/or those with a lot of chest wall fat [11].

The rise of high computational power, the progresses in machine learning (ML) algorithms and the abundance of available data have established artificial intelligence (AI) as a prominent factor in healthcare. ML models have demonstrated remarkable potential in delivering precise diagnostic assessments, identifying appropriate treatment options, and predicting patient outcomes [12]. Several studies have therefore used ML models to diagnose patients with different HCV stages. A recent study by [13] assessed the accuracy of real-time shear wave elastography in identifying liver cirrhosis. The study concluded that this technique reached 92.86% sensitivity, 89.66% specificity, and 91.23% accuracy. However, to the best of our knowledge, no study has focused on developing an ML model to detect the development of cirrhosis in hepatitis C patients. Accordingly, this paper proposes an ML model that detects the presence of cirrhosis in HCV patients to assist medical specialists in recommending timely treatment plans.

Despite the success that ML algorithms have reached in many classification problems, they are still distrusted by end users and people without technical expertise. This is because they have little understanding of how AI models work. To build trust in ML models, researchers have adopted explainable artificial intelligence (XAI) to demonstrate and explain how their models reach a decision [14,15]. This study used Shapley additive explanations (SHAP) and local interpretable model-agnostic explanations (LIME) to explain the outcomes of the best-performing model.

The subsequent sections of this paper are structured as follows. Section 2 comprises a comprehensive literature review. Section 3 outlines the materials and methods employed, encompassing the dataset, the ML algorithms utilized, the performance metrics utilized to assess the proposed models, and the optimization strategy. Section 4 elaborates on the study's findings and the technique used for feature selection. Section 5 describes the models' outcomes using XAI techniques. Finally, Section 6 contains the conclusion and discussion of potential future research avenues.

*Contribution*

Given the gravity of the consequences associated with cirrhosis in individuals with Hepatitis C, it is imperative to reduce unnecessary medical check-ups and optimize time utilization for both medical professionals and patients. Below is a summary of the study's contribution:

- Compare the performance of ensemble learners in diagnosing cirrhosis in hepatitis C patients;
- Apply SFS to minimize the number of features required to form the diagnosis;
- Utilize XAI techniques to explain the outcomes of the best-performing model;
- Utilize XAI techniques to identify the most significant attributes for diagnosing cirrhosis in hepatitis C patients.

## 2. Literature Review

Mostafa et al. [16] used supervised ML algorithms, including an artificial neural network (ANN), a support vector machine (SVM), and an RF, for early diagnosis of hepatitis C. The classifiers were trained using an HCV dataset gathered from the (UCI) ML Repository [17]. The authors found that RF was the best-performing model, achieving an accuracy of 98.14%. Despite the promising results, the model could not be generalized to replace expert knowledge to determine diagnostic paths since the data include numerous missing values.

Similarly, Oladimeji et al. [18] proposed ML models automatically classifying hepatitis C using the same UCI dataset. The authors used several classifiers, including decision tree (DT), RF, k-nearest neighbors (KNN), logistic regression (LR), and naive Bayes (NB). After evaluating all five algorithms, the results indicated that RF outperformed other models with a precision-recall curve of 1.00, an F-measure of 0.99, a Matthews correlation coefficient of 0.99, a receiver operating characteristic area under the curve (ROC-AUC) of 0.99, and an accuracy of 98.97%.

Likewise, Safdari et al. [19] used several classification algorithms to categorize individuals with suspected HCV. Six classification algorithms were used, including SVM, Gaussian naive Bayes (GNB), RF, DT, LR, and KNN, and they were trained using the same UCI dataset. After evaluating the six models according to various measures, the authors found that the RF classifier surpassed the others with an accuracy of 97.29%.

Kaunang [20] also attempted to predict HCV using ML approaches on the UCI dataset. The five categories in the original dataset were reduced to two: the blood donor and suspect blood donor categories were combined into a non-hepatitis category, while the hepatitis, fibrosis, and cirrhosis categories were combined into a hepatitis category. The ML algorithms used were KNN, SVM, RF, ANN, NB, and LR. The LR approach surpassed the other algorithms with an accuracy of 97.9%. However, this study required additional analysis because of a data imbalance between the two classes.

Similarly, Li et al. [21] developed an AI-driven model that has the potential to diagnose HCV and detect the disease at an early stage for potential future treatments. By leveraging the UCI dataset, the researchers used a two-stage cascade strategy that combined the RF and LR algorithms. The artificial bee colony algorithm was utilized to establish the ideal threshold needed for filtering and partitioning. The approach was able to predict the probability of HCV incidence across multiple classes, achieving a high level of accuracy 96.19%, precision 96.94%, recall 96.19%, and F1-score 95.92%.

Ghazal et al. [22] presented an effective and efficient method for assisting healthcare professionals in the early detection of HCV using ML algorithms. A Gaussian SVM model was trained using the Egyptian cohort from the UCI repository [23]. The dataset contained 1385 patient entries, each with 29 distinct attributes. The model achieved an accuracy of 97.9%.

Butt et al. [24] proposed an Intelligent Hepatitis C Stage Diagnosis System that uses an ANN to predict the stage of hepatitis C in a patient using the dataset in [23]. Using 70% of the dataset during training and 30% during validation, the proposed system achieved a precision of 98.89% and 94.44%, respectively.

Mamdouh et al. [25] aimed to detect HCV among healthcare staff in Egypt. Two experiments were conducted, one with feature selection and the other without. The features were chosen using SFS. Then, four algorithms, namely, RF, NB, KNN, and LR, were trained in each experiment. The dataset used for this study was developed at Menoufia University based on records obtained from the National Liver Institute. The dataset included 12 different attributes of 859 participants. It was found that using only four features, RF reached the highest accuracy of 94.88%. However, the dataset was limited to Egyptian patients working in risky environments. In addition, the size of the dataset and the features included were not enough to generalize the model to newly infected patients.

Barakat et al. [26] aimed to build an intelligent diagnostic system using ML to predict and assess fibrosis in children affected with chronic hepatitis C. They used a clinical dataset collected from 166 Egyptian children with this condition. The authors used the RF algorithm to predict the type of fibrosis (no fibrosis, mild, or advanced). The system achieved an accuracy of 87.5% and an AUC-ROC of 90.3%. The prediction of mild fibrosis attained an accuracy of 66% and an AUC of 71%. For advanced fibrosis, it achieved an accuracy of 80% and an AUC of 89.4%. However, the dataset had a limited size and suffered from imbalance.

Similarly, Tsvetkov et al. [27] aimed to develop and test an ML model that detects fibrosis in the liver of individuals with chronic hepatitis C using private data collected from routine clinical examinations. The authors examined data on 1240 patients with chronic hepatitis C, of which 686 were males and 554 were females. A total of 689 patient data were used to develop and test the ML model to obtain the liver fibrosis stage level, and only 9 out of the 28 features were considered. The model attained an accuracy of 80.56%, a sensitivity of 66.67%, and a specificity of 94.44%. Although the study did not employ abnormal or unbalanced samples, did not reject data at random, and tested accuracy with two separate test samples, the ML model still needed external validation.

ML algorithms have had a significant impact in aiding healthcare providers to detect HCV at earlier stages [16]. Detecting individuals with early cirrhosis is critical for preventing severe complications. There have been a few studies that attempted to distinguish the cirrhosis stage from the fibrosis stage. However, there is room for improving the results obtained by their models and reducing their computational complexity using feature selection techniques. Additionally, the outputs of these ML models are not easily understood by medical professionals. Accordingly, in this study, XAI techniques were used to ensure that specialists can understand the model's decisions.

## 3. Materials and Methods

In this study, Python programming language (version 3.9.12) was utilized. First, pre-processing techniques were conducted before building the models. First, binarization focused on predicting only HCV patients with cirrhosis. Consequently, the cirrhosis class was converted to 1 and all other classes were converted to 0. Next, two outliers were identified using a box-plot graphical representation and the interquartile range (IQR) method by comparing the lower bound (first quartile) and upper bound (third quartile) of the data. The values below the first quartile and above the third quartile by 1.5 times the IQR were considered outliers and were removed to improve subsequent analyses or modeling accuracy. Thereafter, the data were normalized using the min-max scaler from the scikit-learn library (version 1.1.1). Due to the conversion of the target class, the data imbalance issue has appeared. Therefore, random oversampling was applied using the imblearn library (version 0.9.1). Subsequently, a stratified k-fold cross-validation approach was used to evaluate four ML algorithms, namely, RF, GBM, XGBoost, and ET, using 10-folds. GridSearchCV from the scikit-learn library was used to tune the algorithms' hyperparameters, and SFS from the mlxtend (version 0.21.0) library was used for feature selection. After building the models, their performance was evaluated and compared using four metrics: accuracy, precision, recall, and AUC-ROC. Lastly, XAI techniques were used to explain the outputs of the best-performing model. Figure 1 illustrates the framework of the study.

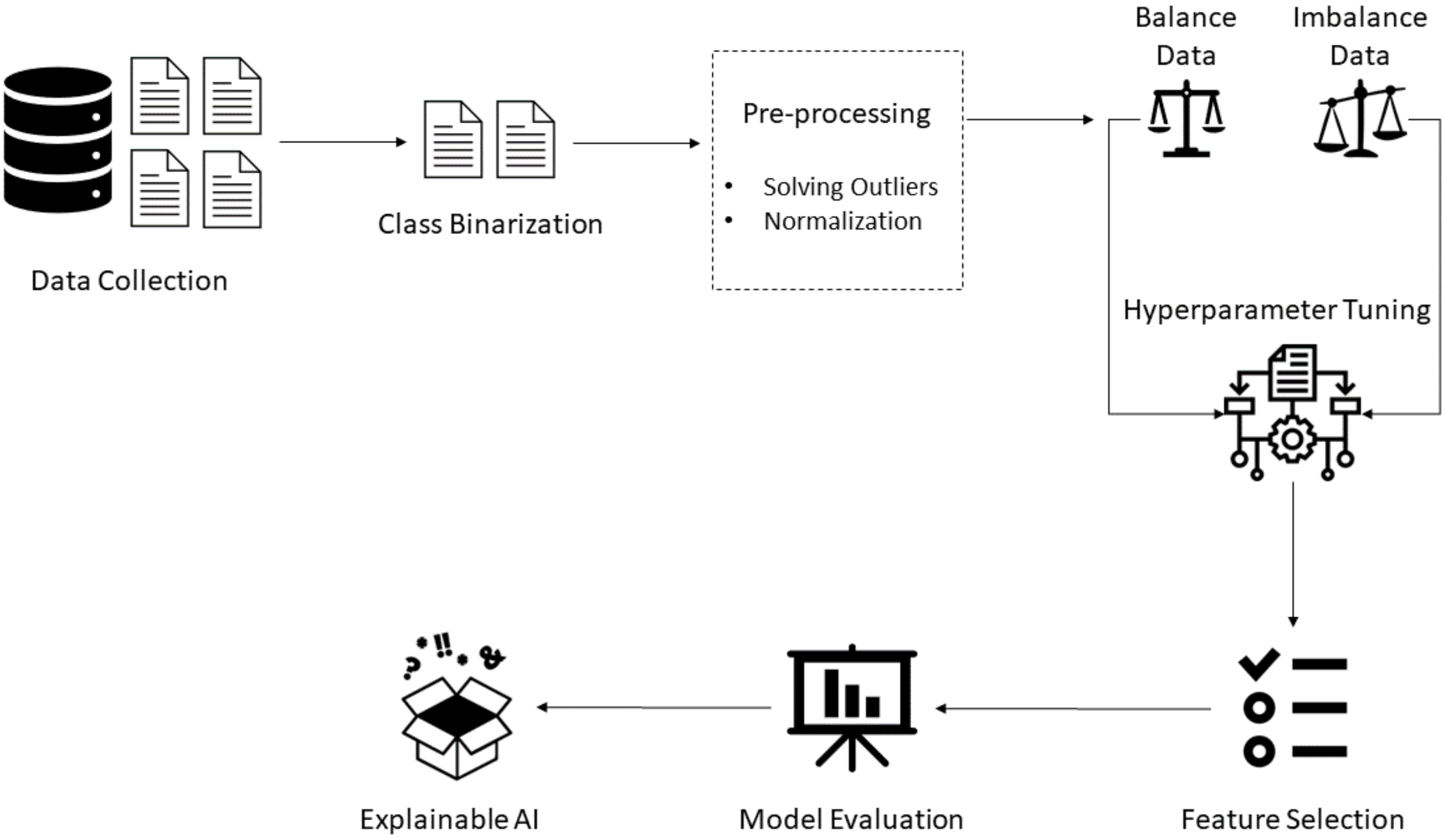

**Figure 1.** Framework of the study.

### *3.1. Dataset Description and Analysis*

The present study is based on the HCV dataset from the UCI ML repository [23]. This dataset includes 29 features of 1385 Egyptian patients, including the target class, who had HCV therapy for approximately 18 months. There are four identifiable stages of hepatitis C virus (HCV) included in the dataset: portal fibrosis without septa, portal fibrosis with a small number of septa, portal fibrosis with many septa, and cirrhosis. Figure 2 shows the sample distribution for each category. More details about the dataset are present in [23,28].

Tables 1 and 2 outline the statistical analysis of the numerical and categorical attributes. The tables show that the dataset has a nearly equal distribution of cases for each categorical feature, which may guarantee the model's generalizability utilizing those features. Moreover, some outliers are indicated from the statistical analysis applied to the numerical features, using the IQR method.

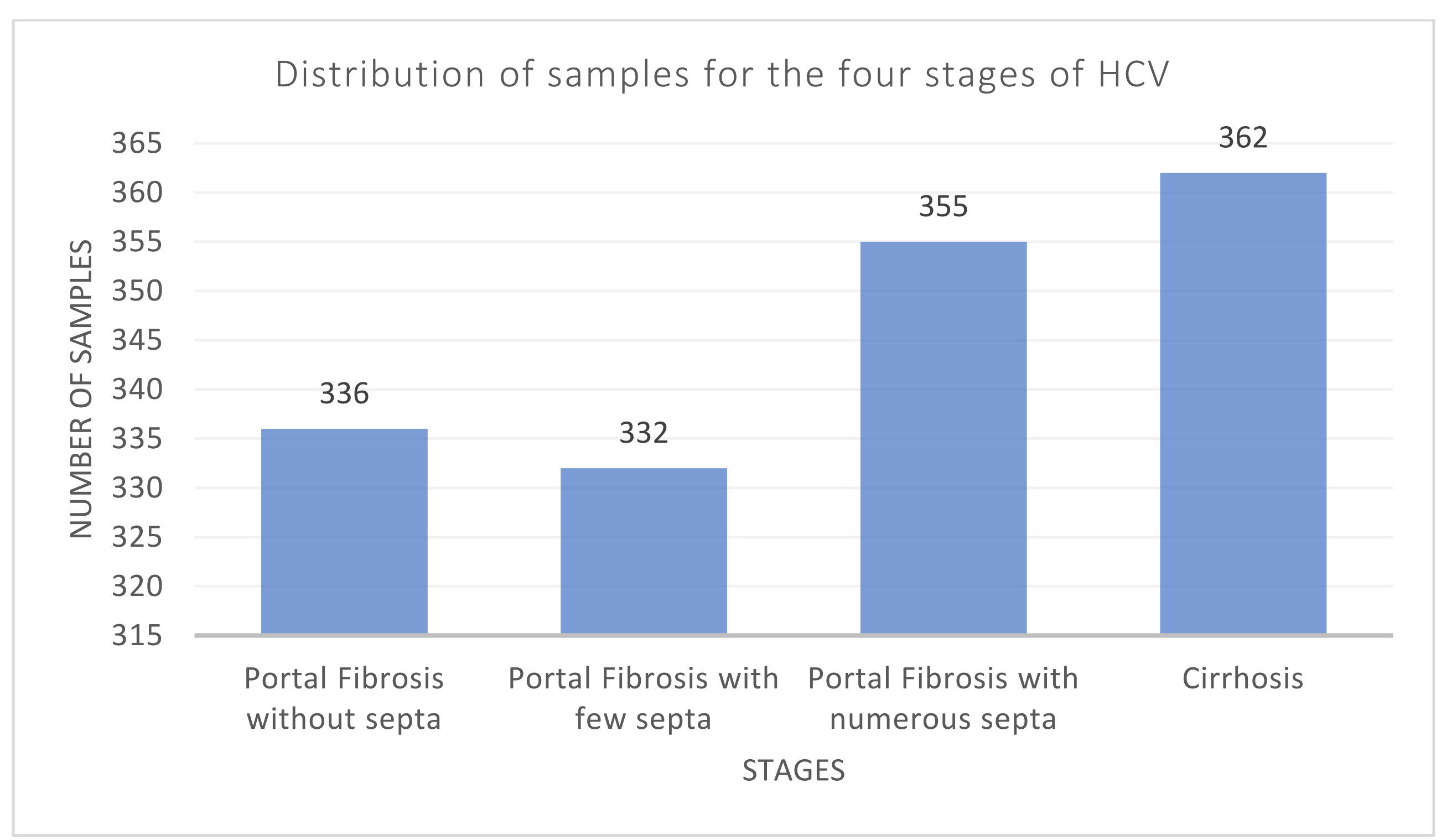

**Figure 2.** Distribution of samples for the four stages of HCV.

**Table 1.** Numerical attributes statistical analysis.

| Feature | Mean | std | min | 25% | 50% | 75% | max |
|---|---|---|---|---|---|---|---|
| Age | 46.32 | 8.78 | 32.00 | 39.00 | 46.00 | 54.00 | 61.00 |
| Body mass index (BMI) | 28.61 | 4.08 | 22.00 | 25.00 | 29.00 | 32.00 | 35.00 |
| White blood cells (WBC) | 7533.39 | 2668.22 | 2991.00 | 5219.00 | 7498.00 | 9902.00 | 12,101.00 |
| Red blood cells (RBC) | 4,422,129.61 | 346,357.71 | 3,816,422.00 | 4,121,374.00 | 4,438,465.00 | 4,721,279.00 | 5,018,451.00 |
| Hemoglobin (HGB) | 12.59 | 1.71 | 10.00 | 11.00 | 13.00 | 14.00 | 15.00 |
| Platelets (Plat) | 158,348.06 | 38,794.79 | 93,013.00 | 124,479.00 | 157,916.00 | 190,314.00 | 226,464.00 |
| Aspartate transaminase ratio (AST 1) | 82.77 | 25.99 | 39.00 | 60.00 | 83.00 | 105.00 | 128.00 |
| Alanine transaminase ratio 1 week (ALT 1) | 83.92 | 25.92 | 39.00 | 62.00 | 83.00 | 106.00 | 128.00 |
| Alanine transaminase ratio 4 week (ALT4) | 83.41 | 26.53 | 39.00 | 61.00 | 82.00 | 107.00 | 128.00 |
| Alanine transaminase ratio 12 week (ALT 12) | 83.51 | 26.06 | 39.00 | 60.00 | 84.00 | 106.00 | 128.00 |
| Alanine transaminase ratio 24 week (ALT 24) | 83.71 | 26.21 | 39.00 | 61.00 | 83.00 | 107.00 | 128.00 |
| Alanine transaminase ratio 36 week (ALT 36) | 83.12 | 26.40 | 5.00 | 61.00 | 84.00 | 106.00 | 128.00 |
| Alanine transaminase ratio 48 week (ALT 48) | 83.63 | 26.22 | 5.00 | 61.00 | 83.00 | 106.00 | 128.00 |
| Alanine transaminase after 24 weeks (ALT after 24 w) | 33.44 | 7.07 | 5.00 | 28.00 | 34.00 | 40.00 | 45.00 |
| RNA Base | 590,951.22 | 353,935.36 | 11.00 | 269,253.00 | 593,103.00 | 886,791.00 | 1,201,086.00 |
| RNA after 4 weeks (RNA4) | 600,895.65 | 362,315.13 | 5.00 | 270,893.00 | 597,869.00 | 909,093.00 | 1,201,715.00 |
| RNA after 12 weeks (RNA12) | 288,753.61 | 285,350.67 | 5.00 | 5.00 | 234,359.00 | 524,819.00 | 3,731,527.00 |
| RNA end-of-treatment (RNAEOT) | 287,660.34 | 264,559.53 | 5.00 | 5.00 | 251,376.00 | 517,806.00 | 808,450.00 |
| RNA elongation factor (RNAEF) | 291,378.29 | 267,700.69 | 5.00 | 5.00 | 244,049.00 | 527,864.00 | 810,333.00 |
| Baseline histological Grading | 9.76 | 4.02 | 3.00 | 6.00 | 10.00 | 13.00 | 16.00 |

**Table 2.** Categorical attributes statistical analysis.

| Feature | Value | Count |
|---|---|---|
| Gender | 1 | 707 |
| | 2 | 678 |
| Fever | 1 | 671 |
| | 2 | 714 |
| Nausea/Vomiting | 1 | 689 |
| | 2 | 696 |
| Headache | 1 | 698 |
| | 2 | 687 |
| Diarrhea | 1 | 689 |
| | 2 | 696 |
| Fatigue and generalized bone ache | 1 | 694 |
| | 2 | 691 |
| Epigastric pain | 1 | 687 |
| | 2 | 698 |
| Jaundice | 1 | 691 |
| | 2 | 694 |

To detect the presence of cirrhosis in hepatitis C patients, patients with stages 1, 2, and 3 were considered negative (portal fibrosis), whereas patients with stage 4 were considered positive (cirrhosis). The random oversampling technique was used to balance the data. Table 3 shows the data before and after outlier removal and random oversampling.

**Table 3.** Number of samples after outlier removal and random oversampling.

| Stage | Number of Samples | Number of Samples after Removing Outliers | Number of Samples after Using Random Oversampling |
|---|---|---|---|
| 0 | 1023 | 1021 | 1021 |
| 1 | 362 | 360 | 1021 |

### *3.2. Description of the Utilized Machine Learning Techniques*

Ensemble algorithms involve training multiple models and combining their results. The bagging classifiers combine several independent predictors using weighted averages or majority votes. In contrast, the boosting classifiers are iterative ensemble methods that modify an observation's weight depending on the most recent classification. If an observation was mistakenly classified, it attempts to enhance the weight of that observation. In this study, two bagging (ET and RF), and two boosting (XGBoost and GBM) techniques were used.

#### 3.2.1. Random Forest

The RF classifier was first introduced by Leo Breiman and Adele Cutler [29]. It is a supervised ML algorithm used in classification and regression problems. An RF consists of an ensemble of many distinct decision trees running parallel as a committee [30]. Incorporating such models improves the performance of the RF classifier, making it more effective than models that operate individually [29]. In classification problems, each decision tree selects a class as an output. The final outcome returned by the RF classifier is produced by taking the highest vote among all trees' outputs. The majority voting formula is:

$$C(x) = \text{mode}\{h_1(x), h_2(x), \ldots, h_n(x)\}, \tag{1}$$

where $C(x)$ represents the predicted class and $h_1(x), h_2(x), \ldots, h_n(x)$ are the $n$ classification models of the data sample $x$ [31].

#### 3.2.2. Gradient Boosting Machine

Leo Breiman first introduced GBM in 1998, where adaptive boosting was defined as a gradient descent with a specific loss function. A year later, Jerome Friedman developed GBMs, which generalize boosting algorithms for regression and classification problems [32]. A common framework for GBM typically involves three fundamental components: an optimized loss function, a weak learner that generates predictions, and an additive model that integrates base learners to decrease the loss function and create a prediction model that is both robust and dependable. Boosting techniques differ from standard ML algorithms because optimization is not included in the function space. However, an optimal function $F(X)$ is reached after $m$ iterations [33],

$$F(X) = \sum_{i=0}^{m} f_i(x), \tag{2}$$

where $f_i(x)$ $(i = 1, \ldots, M)$ represents a feature increment and $f_i(x)$ is calculated using

$$f_i(x) = -p_i g_m(x) \tag{3}$$

where $p_i$ is the loss function and $g_m$ is the negative gradient for the $m$th iteration.

#### 3.2.3. Extreme Gradient Boosting

XGBoost is an ML classifier that employs a combination of gradient boosting and ensembling methods, and it is built upon decision trees as its base learners [34]. XGBoost was initially developed in 2016 by Tianqi Chen and Carlos Guestrin as part of a research project at the University of Washington. The boosting strategy employed by XGBoost involves aggregating multiple models to create a group of predictors that work together to enhance the accuracy of predictions, regardless of whether the problem being addressed is related to classification or regression [34]. The prediction outcome generated by XGBoost is the sum of the scores predicted by the individual decision trees [35],

$$\hat{y} = \sum_{k=1}^{K} f_k(x_i), f_k \epsilon F \tag{4}$$

where $K$ is the number of trees, $f_k(x_i)$ is the score of the $k$th tree, and $F$ is the set of space functions that include all gradient-boosted trees.

XGBoost tackles the issue of overfitting, which can be a considerable concern for ensemble models, by including additional regularization in its objective function. This regularization element punishes the intricacy of the model, enhancing its ability to generalize and decreasing the possibility of overfitting [35]. It is given by:

$$L^{(t)} = \sum_{i} l(y_i, \hat{y}_i) + \sum_{k} \Omega(f_k), \tag{5}$$

The equation involves the loss function $l(y_i, \hat{y}_i)$, which quantifies the difference between the target value $y_i$ and the predicted value $\hat{y}_i$, and the regularization term $\Omega(f_k)$, which evaluates the complexity of the model.

#### 3.2.4. Extra Trees

The extra trees (ET) algorithm operates by picking a subset of features at random and then using them to train a decision tree. After that, the tree is pruned to include only the most valuable features for making predictions. ET is a similar algorithm to RF in that it makes a final prediction about which class or category a data point belongs to by using a collection of decision trees. ET differs from RF because it uses the entire original sample rather than sub-sampling and replacing data as RF does. Another distinction is how the nodes are divided. ET chooses random splits, whereas RF always chooses the best possible split. ET and RF are both designed to improve the final output [36]. Decision trees, RF and ET also differ in performance: the variance is high in decision trees, medium in RF, and low in ET [37].

### 3.3. Performance Measures

Four measures assessed classification performance: accuracy, precision, recall, and AUC-ROC. In addition, to assess the performance of the models, a confusion matrix was formed for each model, which evaluates their true positives (TP), false positives (FP), false negatives (FN), and true negatives (TN). TP represents the number of correctly classified patients with cirrhosis-HCV, while FP represents the number of patients incorrectly classified as cirrhosis-HCV. FN is the number of patients incorrectly classified as non-cirrhosis-HCV, and TN is the number of correctly classified non-cirrhosis-HCV patients.

Accuracy is the ratio of accurately classified observations to total observations,

$$\text{Accuracy} = \frac{TP + TN}{TP + FP + FN + TN}. \tag{6}$$

Precision is the ratio of accurately classified positive observations to the total number of positively classified observations,

$$\text{Precision} = \frac{TP}{TP + FP}. \tag{7}$$

Recall is the ratio of accurately predicted positive observations in a class to the total number of observations,

$$\text{Recall} = \frac{TP}{TP + FN}. \tag{8}$$

### 3.4. Optimization Strategy

The hyperparameters of an algorithm must be modified in order to generate models that can solve problems optimally. Grid search with stratified 10-fold cross-validation was utilized in this study for this purpose. Grid search is used to define a search space by specifying the hyperparameters and their range of potential values. After defining the hyperparameter grid, the GridsearchCV technique generates every possible combination of the values to identify the optimal set of hyperparameters. To assess the efficacy of each hyperparameter combination, 10-fold cross-validation is conducted to evaluate the performance of the model. Table 4 highlights the optimal hyperparameters for models using the datasets before and after oversampling.

**Table 4.** The optimal hyperparameters for each classifier using original and oversampled data.

| Classifier | Hyperparameter | Without Oversampling | With Oversampling |
|---|---|---|---|
| RF | n_estimators | 100 | 450 |
| | max_depth | 23 | 24 |
| | max_features | sqrt | log2 |
| | max_samples_leaf | 1 | 1 |
| GBM | n_estimators | 50 | 110 |
| | learning_rate | 0.01 | 0.3 |
| | max_depth | 1 | 10 |
| | loss | log_loss | exponential |

**Table 4.** *Cont.*

| Classifier | Hyperparameter | Without Oversampling | With Oversampling |
|---|---|---|---|
| XGBoost | n_estimators | 50 | 170 |
| | booster | gblinear | gbtree |
| | learning_rate | 0.01 | 0.1 |
| | gamma | 0 | 0.4 |
| ET | n_estimators | 50 | 200 |
| | max_depth | None | 11 |
| | max_features | log2 | log2 |
| | min_samples_leaf | 1 | 1 |

## 4. Results and Discussion

This section presents the proposed models' outcomes after applying the GridsearchCV method to the models, using both the original and sampled data. Table 5 assesses the performance of the constructed models through stratified 10-fold cross-validation.

**Table 5.** The results of the proposed models before and after random oversampling was applied.

| Classifier | Dataset | Mean of Accuracy | Std of Accuracy | Precision | Recall | AUC-ROC |
|---|---|---|---|---|---|---|
| RF | Original | 74.22% | 0.0047 | 3.00% | 1.00% | 0.49 |
| | Oversampled | 96.48% | 0.0417 | 99.27% | 93.64% | 0.96 |
| GBM | Original | 73.93% | 0.0006 | 0.00% | 0.00% | 0.45 |
| | Oversampled | 95.70% | 0.0439 | 97.49% | 93.74% | 0.97 |
| XGBoost | Original | 73.93% | 0.0006 | 0.00% | 0.00% | 0.52 |
| | Oversampled | 90.99% | 0.0396 | 88.00% | 94.81% | 0.96 |
| ET | Original | 74.22% | 0.0060 | 45.00% | 1.66% | 0.51 |
| | Oversampled | 96.82% | 0.0413 | 100% | 93.64% | 0.97 |

The results demonstrate significant differences in the accuracy, precision, and recall before and after applying the random oversampling algorithm. It is important to consider the increase in precision and recall rates, which concentrate on the number of FP and FN, since cirrhosis is a chronic disease that must be detected early to prevent complications. Table 5 shows that the boosting models failed to predict any positive cases without oversampling, attaining a recall and precision rate of 0%. ET produced the best results before random oversampling, with an accuracy of 74.22%, a precision of 45%, and a recall of 1.6%. It also attained the highest accuracy of 96.82% after the random oversampling. Although random oversampling is known to increase the possibility of oversampling, it considerably improved the performance of all models. This method may be useful for ML algorithms that are impacted by skewed distributions and where the model fit may be influenced by several duplicate samples for a particular class. This may include algorithms that iteratively learn coefficients and that seek good splits of data, such as decision trees. In the subsequent experiments, the models trained on the oversampled data were used.

### *4.1. Feature Selection*

Feature selection is critical to developing efficient models since it eliminates irrelevant features that might negatively influence their performance. The SFS technique from the mlxtend library was used to reduce the number of features [38]. This algorithm trains a model using the optimal features selected by a particular criterion function. The SFS algorithm selects the feature that enhances the chosen criterion function, which is included in the feature subset in every subsequent forward step. The most suitable feature subset identified by the SFS algorithm is presented in Table 6. As can be seen in this table, the SFS algorithm did not affect the performance of the ensemble boosting models. However, there was a slight reduction in features for the RF model and a significant reduction for the ET model. The accuracy of both classifiers was improved by 0.1% by SFS, despite a considerable difference in the number of removed attributes.

**Table 6.** SFS results.

| Classifier | Number of Features Selected | Features Selected | Mean of Accuracy |
|---|---|---|---|
| RF | 27 | {Jaundice, Age, Gender, Fatigue & generalized bone ache, ALT 48, RNA 48, RNA base, BMI, HGB, ALT 12, RNA EOT, ALT 4, Nausea/Vomiting, ALT 1, Epigastric pain, Fever, Plat, AST 1, RNA EF, Headache, Baseline histological grading, ALT 24, RBC, ALT after 24 w, WBC, ALT 36, RNA 4, Diarrhea} | 96.58% |
| GBM | 28 | All features | 95.70% |
| XGBoost | 28 | All features | 90.99% |
| ET | 16 | {Age, BMI, Diarrhea, Jaundice, RBC, HGB, Plat, AST 1, ALT 12, ALT 24, ALT 36, ALT 48, ALT after 24 w, RNA base, RNA 4, RNA 12} | 96.92% |

### 4.2. Further Discussion of the Results

After analyzing and applying SFS to all proposed models, it has been shown that ET surpassed all other algorithms, achieving 96.92% accuracy, as shown in Table 7, followed by RF, with 96.58% accuracy. XGBoost achieved the lowest accuracy of 90.99%; however, it achieved the highest recall of 94.81%. ET obtained the highest precision of 99.81%. To further evaluate the results in terms of TP, FP, FN, and TN counts, confusion matrices were constructed. These are displayed in Figure 3.

**Table 7.** The performance of the models after SFS.

| Classifier | Mean of Accuracy | Std of Accuracy | Recall | Precision |
|---|---|---|---|---|
| RF | 96.58% | 0.0399 | 93.74% | 99.39% |
| GBM | 95.70% | 0.0439 | 93.74% | 97.49% |
| XGBoost | 90.99% | 0.0396 | 94.81% | 88.00% |
| ET | 96.92% | 0.0380 | 94.00% | 99.81% |

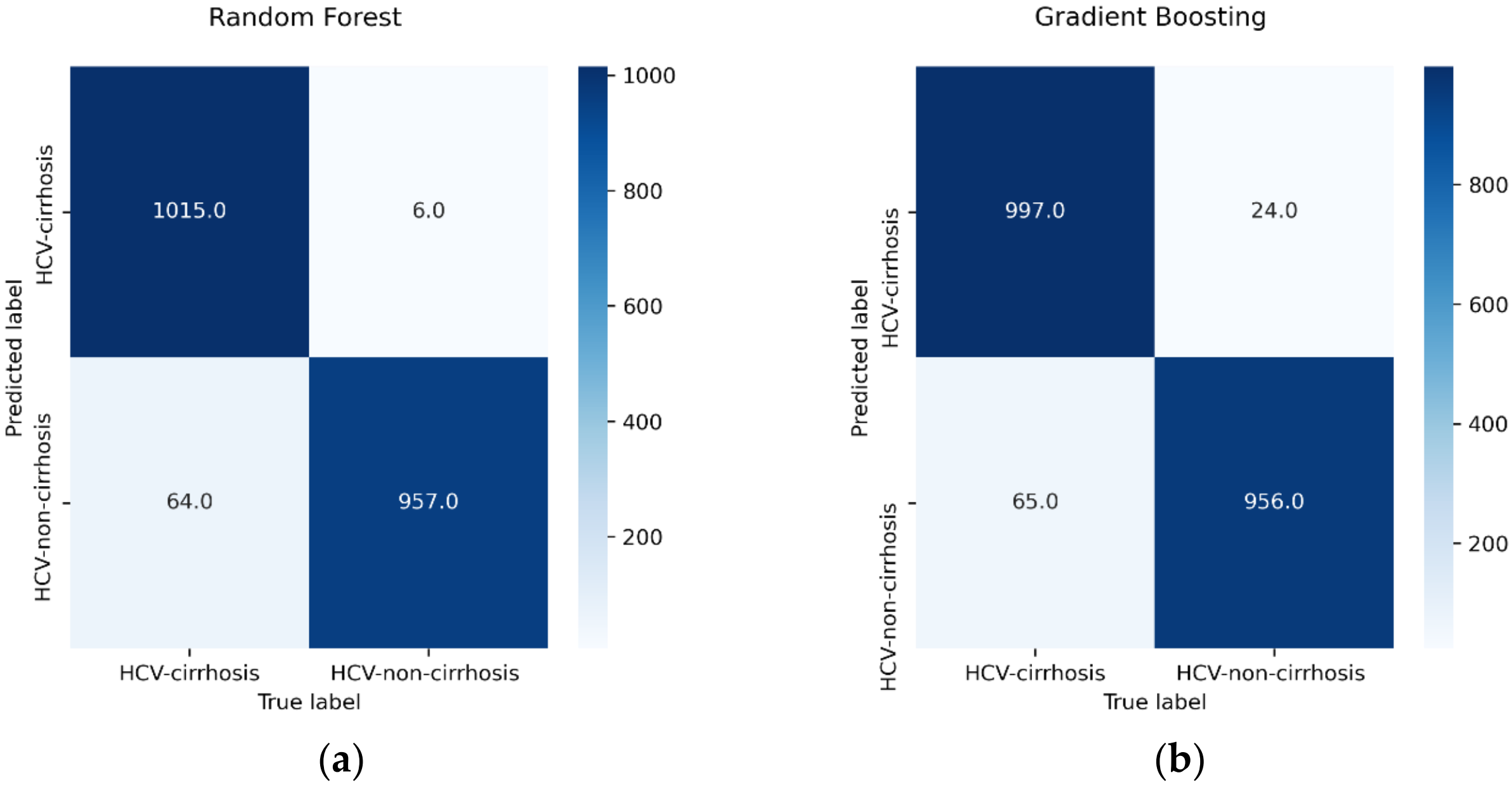

**Figure 3.** *Cont.*

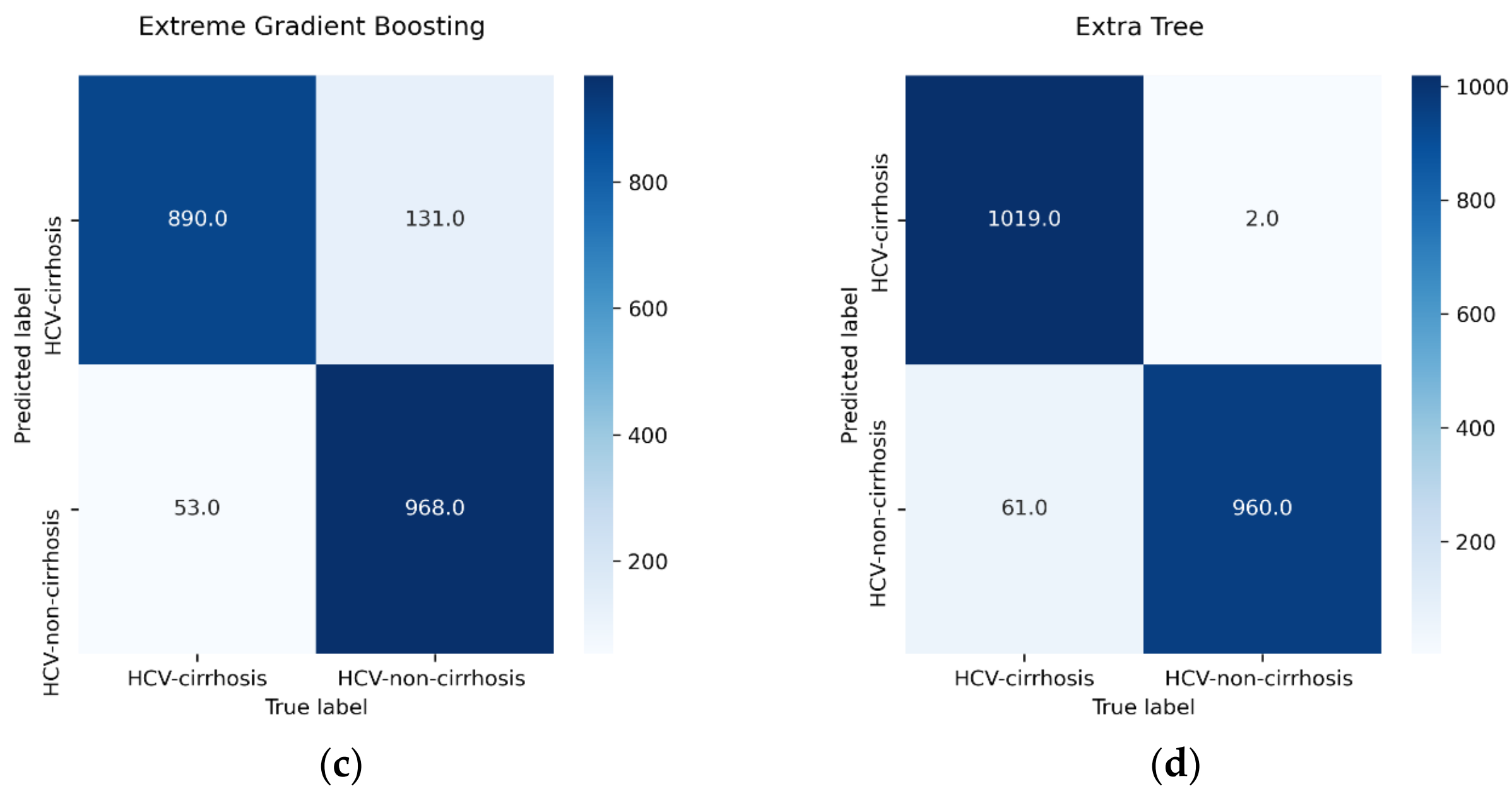

**Figure 3.** Confusion matrices: (**a**) RF, (**b**) GBM, (**c**) XGBoost, (**d**) ET.

According to the findings illustrated in Figure 3, XGBoost displayed the lowest count of FN, followed by ET. On the other hand, ET demonstrated the lowest number of FP, with RF ranking second. RF and GBM obtained the highest number of FNs. Meanwhile, XGBoost had the highest FP count of 131. FNs have a significant impact on healthcare, but minimizing the number of FPs is also essential to avoid unnecessary interventions. Misdiagnosis of cirrhosis wastes medicine and time, and it damages patients' mental health. Consequently, to determine the best-performing model, the tradeoff between FPs and FNs should be considered. Overall, ET surpassed other models in detecting cirrhosis in HCV patients.

The study's objective was to evaluate the models' capability to differentiate between patients with cirrhosis and those who did not have the condition, using stratified 10-fold cross-validation to construct an AUC-ROC curve for each model. To assess the ability of the models to differentiate between patients with cirrhosis and without, an AUC-ROC curve was constructed for each model using stratified 10-fold cross-validation. The results, presented in Figure 4, indicate how effectively the proposed models can distinguish between different classes to generate correct predictions. The results indicated that all models obtained an outstanding AUC-ROC of 96%.

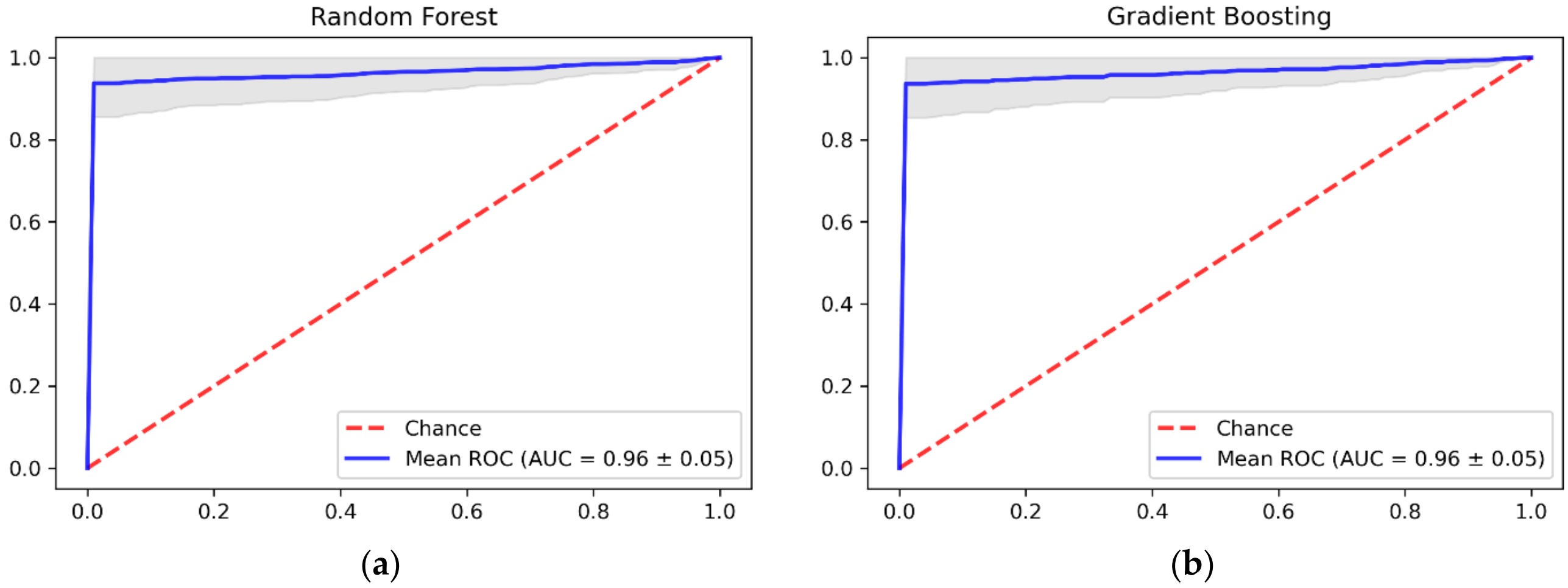

**Figure 4.** *Cont.*

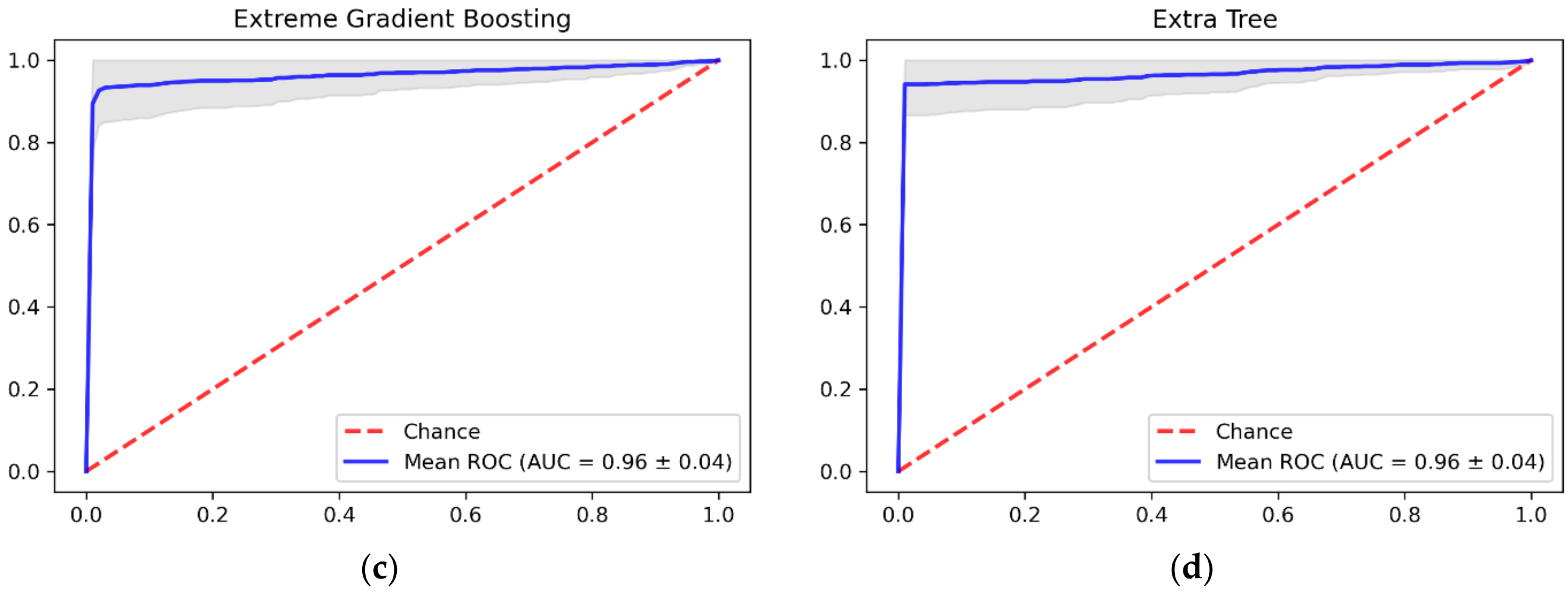

**Figure 4.** AUC-ROC values (**a**) RF, (**b**) GBM, (**c**) XGBoost, (**d**) ET.

## 5. Explainable Artificial Intelligence

ML has gained widespread popularity and has been applied to numerous domains and use cases. However, specific measures need to be implemented to ensure that society accepts and trusts ML-powered systems. To build this trust, it is necessary to visualize and explain how ML models make their decisions. XAI can be used to ensure that an algorithm's decision-making process and the data used for training are easily understood by users. This study uses two XAI techniques, SHAP and LIME.

### *5.1. Shapley Additive Explanations*

In 2017, Lundberg and Lee introduced the SHAP algorithm, which utilizes an additive feature attribution method with a linear model to calculate the contribution of each attribute to a prediction. Specifically, SHAP presents the Shapley value explanation, which allows for the prediction of an instance. The SHAP explanation approach employs coalitional game theory to calculate the Shapley values. In this method, the feature values of data instances are treated as individual actors participating in a coalition. By treating the feature values of a data instance as actors in the coalition, researchers can use Shapley values to distribute the prediction equally across the features. The explanation is calculated using:

$$g(z') = \varnothing_0 + \sum_{j=1}^{M} \varnothing_j z'_j, \quad (9)$$

This equation involves the explanatory model, $g$, the coalition vector, $z' \in \{0,1\}^M$, where $M$ represents the maximum coalition size. The feature attribution for a given feature, $j$, is represented by the Shapley value $\varnothing_j \in \mathbb{R}$. The Shapley values of the ET model are demonstrated in Figure 5, where positive contributions are displayed on the left side and negative contributions are displayed on the right side [39].

Based on the observations made in Figure 5, it is evident that RNA 4, BMI, RNA 12, and AST 1 are the features with the greatest importance. Conversely, the features with the lowest importance are jaundice and diarrhea. It can also be observed that high RNA 4 values have low negative contributions, while the opposite is true for patients with high RNA 12 values. On the other hand, high BMI and AST 1 values have high positive contributions.

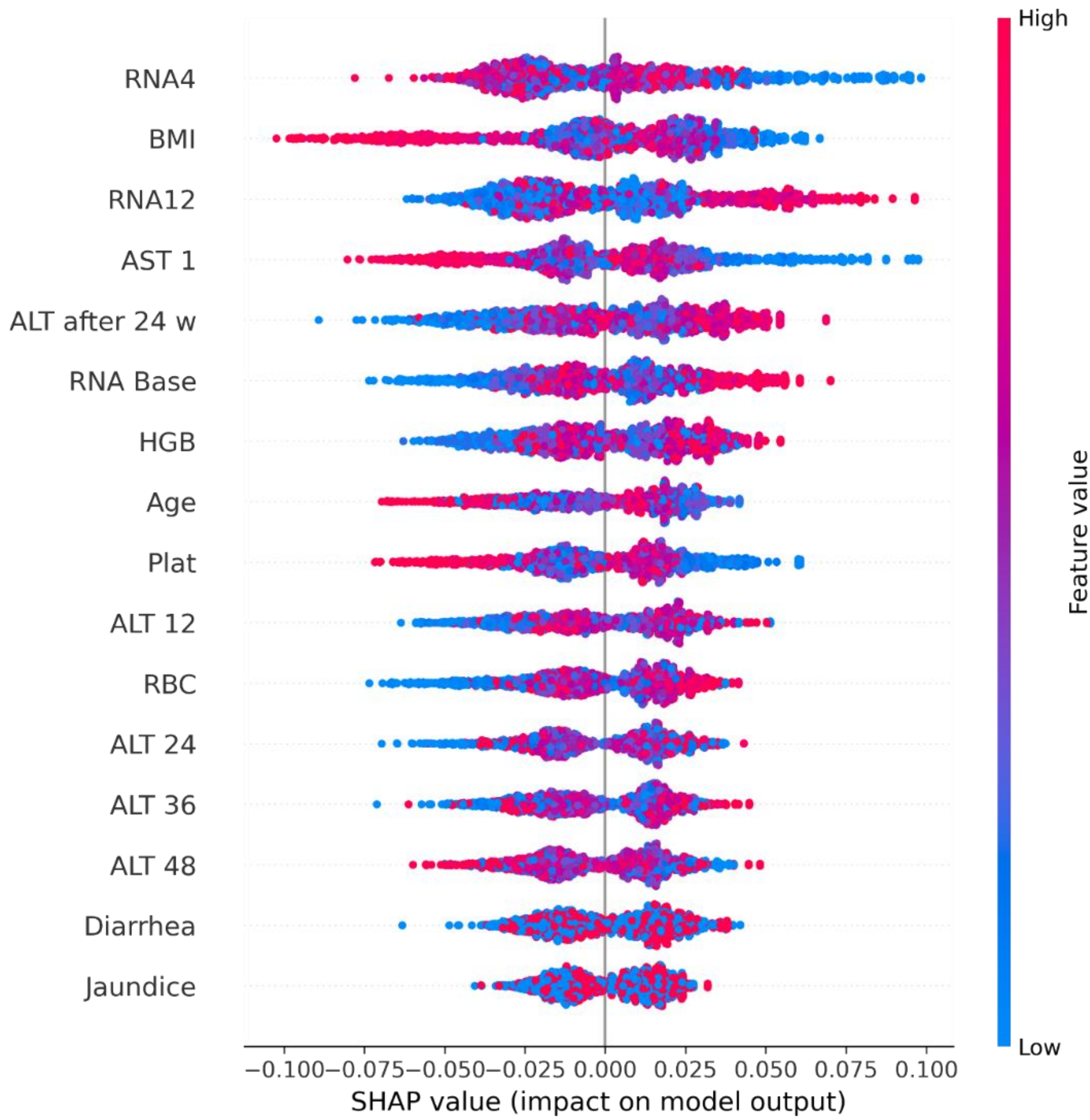

**Figure 5.** Shapley values for the ET model.

*5.2. Local Interpretable Model-Agnostic Explanations*

LIME is a commonly used algorithm that provides the ability to interpret machine learning models by creating a comprehensive explanation for a single prediction. LIME's prediction is based on a simpler interpretable model, such as a linear classifier. In this technique, random perturbation is used to simulate data around an instance, and specific selection techniques are used to determine the importance of certain features. The popularity of LIME and similar local algorithms can be attributed to its ease of use. Still, the generated explanations are unstable because of the random perturbation and feature selection approaches, which can produce many explanations for the same prediction [40]. Figure 6 illustrates the results of LIME for a positive and a negative prediction.

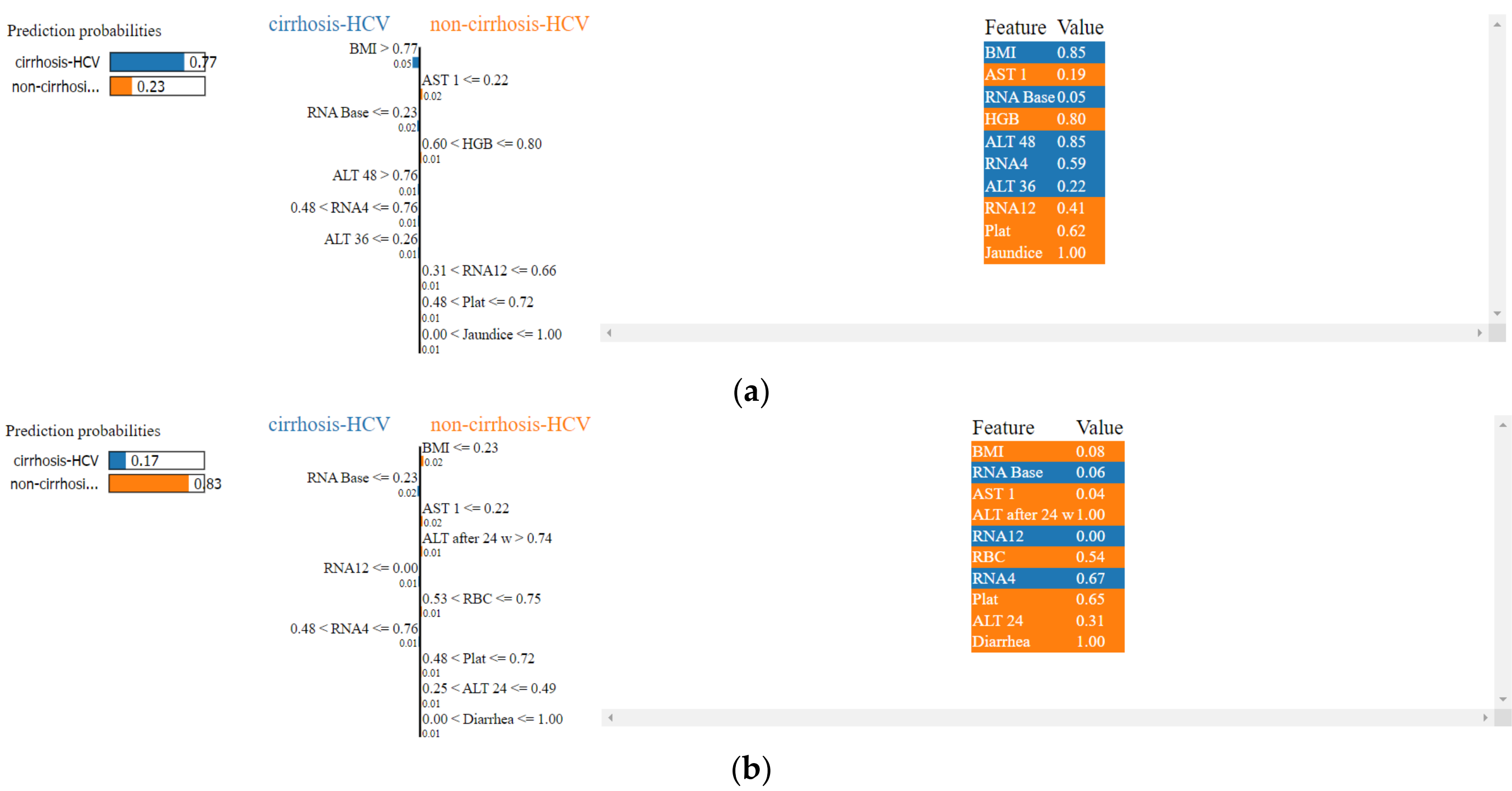

**Figure 6.** LIME predictions for the ET model (**a**) positive, (**b**) negative.

Figure 6a explains a positive prediction made by the ET model, in which the positive probability prediction was 77%. It is indicated that the BMI, RNA Base, ALT 48, RNA 4, and ALT36 contributed to the positive prediction. On the other hand, Figure 6b explains a negative prediction made by the ET model, in which the negative probability prediction was 83%. It can be observed that all features except RNA Base, RNA 12, and RNA 4 contributed to the negative prediction.

## 6. Conclusions and Recommendations

Cirrhosis, caused by extensive liver fibrosis or scarring, is frequently discovered after decompensation when its associated consequences have occurred. The performance of current non-invasive testing for the early detection of advanced liver cirrhosis is poor, with many categories being uncertain. Healthcare professionals can detect the presence of cirrhosis and chronic liver diseases using invasive tests, including liver biopsy. However, ML algorithms can be applied to analyze clinical data to detect the presence of cirrhosis to assist healthcare providers. This study aimed to use ML algorithms to identify cirrhosis in HCV patients. Four algorithms were trained using the Egyptian HCV patient's dataset from UCI, namely, RF, GBM, XGBoost, and ET. The ET classier outperformed the other algorithms using only 16 out of 29 features, with an accuracy of 96.92%, a recall of 94.00%, a precision of 99.81%, and an AUC-ROC of 96%. Although XGBoost achieved a higher recall value, ET had the highest accuracy value. This would result in less clinical testing, possibly contributing to cost savings.

In addition, the use of XAI was implemented in order to guarantee that healthcare experts could comprehend how the algorithm makes decisions and the information utilized to train it. The results of SHAP revealed that the features with the highest importance were RNA 4, BMI, RNA 12, and AST 1. On the other hand, LIME indicated that BMI, RNA Base, ALT 48, RNA 4, and ALT36 contributed to the positive predictions of the model. The results of this study were promising, but more data on patients with cirrhosis are needed to train the model on real data rather than synthetic data. Furthermore, other stages of HCV could be investigated in the future to develop a generalized model that can predict HCV progression earlier.

**Author Contributions:** Conceptualization, A.A. (Abrar Alotaibi), L.A., N.A., A.A. (Alhatoon Alanazy), S.A. (Salam Alshammasi), M.A., S.A. (Shoog Alrassan), and A.A. (Aisha Alansari); methodology, A.A. (Abrar Alotaibi), L.A., N.A., A.A. (Alhatoon Alanazy), S.A. (Salam Alshammasi), M.A., S.A. (Shoog Alrassan) and A.A. (Aisha Alansari); software, L.A., N.A. and A.A. (Aisha Alansari); validation, A.A. (Abrar Alotaibi), L.A., N.A., A.A. (Alhatoon Alanazy), S.A. (Salam Alshammasi), M.A., S.A. (Shoog Alrassan) and A.A. (Aisha Alansari); formal analysis, A.A. (Abrar Alotaibi), L.A., N.A., A.A. (Alhatoon Alanazy), S.A. (Salam Alshammasi), M.A., S.A. (Shoog Alrassan), and A.A. (Aisha Alansari); investigation, A.A. (Abrar Alotaibi), L.A., N.A., A.A. (Alhatoon Alanazy), S.A. (Salam Alshammasi), M.A., S.A. (Shoog Alrassan) and A.A. (Aisha Alansari); resources A.A. (Abrar Alotaibi), L.A., N.A., A.A. (Alhatoon Alanazy), S.A. (Salam Alshammasi), M.A., S.A. (Shoog Alrassan), and A.A. (Aisha Alansari); data curation L.A., N.A., A.A. (Alhatoon Alanazy), S.A. (Salam Alshammasi), M.A. and S.A. (Shoog Alrassan); writing—original draft preparation, L.A., N.A., A.A. (Alhatoon Alanazy), S.A. (Salam Alshammasi), M.A. and S.A. (Shoog Alrassan); writing—review and editing, A.A. (Abrar Alotaibi) and A.A. (Aisha Alansari); visualization, A.A. (Abrar Alotaibi), L.A., N.A., A.A. (Alhatoon Alanazy), S.A. (Salam Alshammasi), M.A., S.A. (Shoog Alrassan) and A.A. (Aisha Alansari); supervision, A.A. (Abrar Alotaibi) and A.A. (Aisha Alansari); project administration, A.A. (Abrar Alotaibi) and A.A. (Aisha Alansari); funding acquisition, A.A. (Abrar Alotaibi), L.A., N.A., A.A. (Alhatoon Alanazy), S.A. (Salam Alshammasi), M.A., S.A. (Shoog Alrassan), and A.A. (Aisha Alansari). All authors have read and agreed to the published version of the manuscript.

**Funding:** This research received no external funding.

**Data Availability Statement:** The data used to support this study and are available at https://archive.ics.uci.edu/ml/datasets/Hepatitis+C+Virus+%28HCV%29+for+Egyptian+patients (accessed on 9 May 2023).

**Conflicts of Interest:** The authors declare no conflict of interest.